\documentclass[letterpaper, 10 pt, conference]{ieeeconf}  % Comment this line out if you need a4paper

\usepackage{graphicx}
\usepackage{cite}
\usepackage{balance} 
\usepackage{upgreek}
\usepackage{multirow}
\usepackage{booktabs}
\usepackage{bbm}
\usepackage{color,soul}
\sethlcolor{yellow}
\usepackage{array}
\usepackage{longtable}
\usepackage{lipsum}
\usepackage{verbatim}
\let\labelindent\relax
\usepackage{enumitem}
\usepackage{caption}
\usepackage{float} 
\usepackage{subfigure}
\usepackage{subcaption}
\usepackage{amsmath}
\usepackage{amssymb}
\usepackage{tabularx}
\usepackage{algorithm}
\usepackage{ragged2e}
\usepackage{lipsum} 
\usepackage{pifont} 
\usepackage{utfsym}

\IEEEoverridecommandlockouts                              % This command is only needed if 
\title{\LARGE \bf
Generalizable 6D Pose Estimation of Textureless Objects with Planar-based Gaussian Splatting 
} 

\author{
Jie Lu$^{1}$, Hengtan Zhang$^{1}$, Li Gong$^{1}$, Pengpeng Wang$^{2}$,
Xianjia Yu$^{3}$, Jinxiang Deng$^{4}$,\\
Tomi Westerlund$^{3}$, Zhongxue Gan$^{2}$,
Lirong Zheng$^{1}$, and Zhuo Zou$^{1}$%
\thanks{*This work was supported in part by the National Key Research and Development Program of China under Grant 2023YFB4704100.}%
\thanks{$^{1}$College of Future Information Technology, Fudan University, Shanghai, China.
{\tt\small jlu24@m.fudan.edu.cn}}%
\thanks{$^{2}$College of Intelligent Robotics and Advanced Manufacturing, Fudan University, Shanghai, China.}%
\thanks{$^{3}$Turku Intelligent Embedded and Robotic Systems (TIERS) Lab, University of Turku, 20014 Turku, Finland.}%
\thanks{$^{4}$Jihua Laboratory, Guangdong, China.}%
}

\begin{document}

\maketitle
\thispagestyle{empty}
\pagestyle{empty}

%%%%%%%%%%%%%%%%%%%%%%%%%%%%%%%%%%%%%%%%%%%%%%%%%%%%%%%%%%%%%%%%%%%%%%%%%%%%%%%%
\begin{abstract}

Estimating the 6D pose of textureless objects without prior CAD models remains a critical challenge due to the lack of appearance features. While recent generalizable approaches alleviate the dependence on object-specific models, their performance on low-texture objects is often limited by insufficient geometric constraints in the underlying representations. In this work, we propose PG-Pose, a geometry-aware framework combining \underline{\textbf{P}}lanar-based Gaussian Splatting (PGS) reconstruction and \underline{\textbf{G}}eometry-driven pose optimization. In the offline representation extraction stage, three distinct representations of the object are extracted from multi-view reference RGB images with known poses. PG-Pose reconstructs a 3D Gaussian representation and renders high-fidelity depth maps to generate 3D point clouds through back projection. In the online pose inference stage, the initial pose of the input image is estimated by 2D-3D correspondence matching between the input image and the reconstructed 3D point clouds, followed by a PGS-Refiner for iterative pose optimization. Evaluations on the OnePose-LowTexture datasets, PG-Pose achieves an average accuracy of 94.2\% ADD(S)@0.1d, with a 2.1\% improvement average accuracy compared with the state-of-the-art (SOTA) GS-based approach. To further demonstrate the effectiveness of PG-Pose for industrial robots in grasping tasks, we deploy it on a dual-arm industrial robot and successfully realize the grasping task on an unseen object.

\end{abstract}

%%%%%%%%%%%%%%%%%%%%%%%%%%%%%%%%%%%%%%%%%%%%%%%%%%%%%%%%%%%%%%%%%%%%%%%%%%%%%%%%

\begin{figure*}[!t]
\centering
    \centerline{\includegraphics[width=0.95\textwidth]{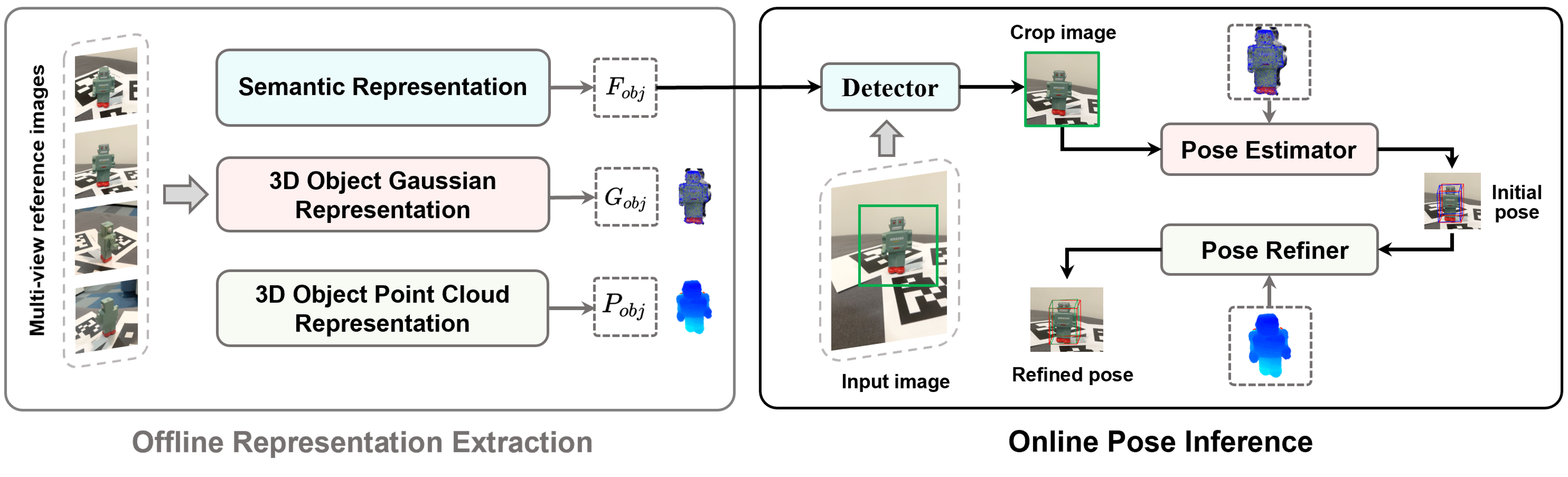}}
\caption{Overall architecture of the proposed 6D pose estimation framework.}
\label{arch}
\end{figure*}

% \begin{figure*}[!t]
% \centering
% \includegraphics[width=\textwidth,height=0.8\textheight,keepaspectratio]{arch.png}
% \caption{Overall architecture of the proposed two-stage framework.}
% \label{arch}
% \end{figure*}

\section{INTRODUCTION}

\par 6D pose estimation aims to acquire the translation and rotation of a target object in 3D space. Accurate and robust algorithms for 6D object pose estimation are crucial in many robotic applications, such as grasping tasks and robotic manipulation\cite{wang2023gsnet}\cite{zhao2024tooleenet}\cite{cavallari2019real}\cite{hodan2020epos}\cite{huang2024sd}.

\par Object-specific pose estimation methods \cite{peng2019pvnet}\cite{Li_Wang_Ji_2019}\cite{Xiang_Schmidt_Narayanan_Fox_2018} achieve high accuracy by training on pre-defined objects with CAD models or dense textures. However, these methods rely on specific instance data, which results in poor generalization to unseen objects. In contrast, category-level approaches\cite{Wang_Sridhar_Huang_Valentin_Song_Guibas_2019}\cite{Chen_Jia_Chang_Duan_Shen_Leonardis_2021} leverage shape priors within a category to handle new instances, but their accuracy degrades when faced with significant intra-category variations in shape and scale, especially for symmetric or geometrically ambiguous objects. 

\par Recently, several studies have made considerable progress in generalizable object pose estimation, aiming to eliminate the dependence on CAD models or instance-specific training. Methods such as Gen6D \cite{liu2022gen6d} utilizes reference RGB images for pose initialization and refinement for pose initialization and refinement, thereby eliminating the need for CAD models. However, Gen6D heavily relies on accurate 2D bounding box detection and struggles with occlusions due to its template-matching mechanism. OnePose \cite{sun2022onepose} and OnePose++ \cite{he2022onepose++} further extend the generalizability by reconstructing sparse 3D point clouds from multi-view RGB inputs and solving poses by 2D-3D correspondences. Although OnePose++ reduces the reliance on key points for low-texture objects through detector-free feature matching, its sparse point cloud representation leads to instability in pose estimation due to geometric ambiguity in symmetric or textureless regions. 

\par To address these limitations, GS-Pose \cite{cai2024gs} pioneeringly introduces a framework based on 3D Gaussian Splatting (3DGS). By building a 3D Gaussian representation of the object and refining pose through differentiable rendering, GS-Pose achieves state-of-the-art (SOTA) performance. Nevertheless, there are two problems with GS-Pose, one is that GS-Pose uses a retrieval method to estimate the initial 6D pose of the object, resulting in a low precision of the initial pose. The other is that its reliance on photometric consistency for Gaussian parameter optimization remains vulnerable to textureless surfaces.

\par To address these issues, we propose PG-Pose, a framework that focuses on enhancing the geometric expressiveness of 3D Gaussian representations for model-free 6D pose estimation. Instead of introducing a new pose estimation pipeline, PG-Pose operates within the widely adopted coarse-to-fine paradigm and improves its robustness by injecting Planar Gaussian priors into the reconstruction process and integrating Geometry-driven initialization for online inference. In the offline stage, the proposed method reconstructs a planar-based 3D Gaussian representation from multi-view RGB images, where geometric regularization terms such as surface normal consistency and coplanarity constraints are integrated into the Gaussian parameter optimization. This results in a geometrically enhanced representation that enables the rendering of high-fidelity depth maps, which are back-projected to generate dense 3D point clouds with improved structural completeness compared to photometric-only reconstruction, particularly under low-texture conditions.

\par In the online stage, benefiting from the improved geometric fidelity of the reconstructed representation, PG-Pose naturally supports explicit 2D-3D correspondence matching and enables a Perspective-n-Point (PnP) based pose initialization. Compared to retrieval-based initialization, this geometry-driven strategy provides a more stable starting point for symmetric and textureless objects. The initial pose is further refined using a PGS-Refiner module based on differentiable Gaussian rendering. Experimental results on textureless benchmark objects as well as deployment on a dual-arm industrial robot demonstrate that PG-Pose achieves superior accuracy and robustness compared to existing CAD-free approaches. The main contributions are listed as follows: 

\begin{itemize}[leftmargin = 10pt]
    \item We propose PG-Pose, a robust 6D pose estimation framework driven by a geometrically enhanced 3D Gaussian representation that integrates planar priors, surface normal consistency, and coplanarity regularization into the Gaussian Splatting optimization process. This design substantially improves reconstruction robustness without relying on CAD models.
    \item Built upon the enhanced representation, we demonstrate that geometry-driven pose initialization and Gaussian-based refinement can be reliably combined within a standard coarse-to-fine framework. PG-Pose achieves 94.2\% ADD(S)@0.1d on the OnePose-LowTexture dataset and robust real-world robotic grasping of unseen objects. Ablation studies further confirm that both planar priors and geometry-driven initialization are critical for handling textureless and symmetric objects.
\end{itemize}

\begin{figure*}[!t]
\centering
    \centerline{\includegraphics[width=0.95\textwidth]{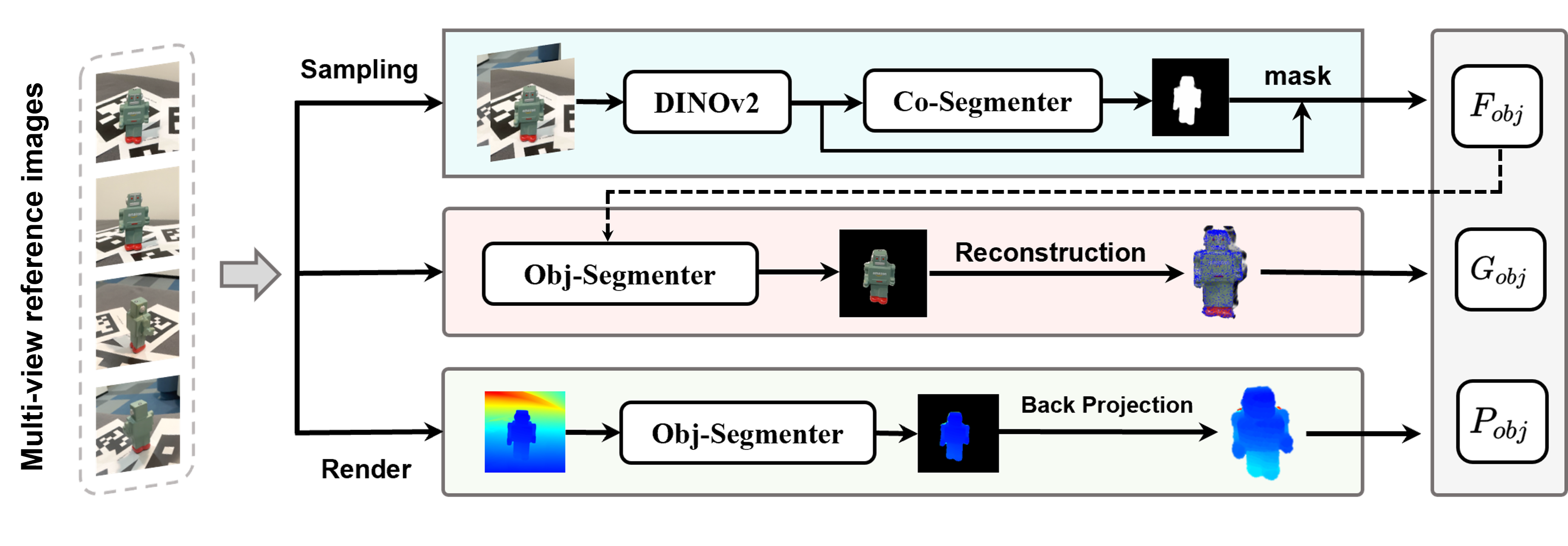}}
\caption{Overview of the representation extraction process.}
\label{database}
\end{figure*}

\section{RELATED WORK}
\subsection{Textureless Pose Estimation}

\par The evolution of 6D pose estimation for textureless objects has been driven by the need to overcome the absence of discriminative visual cues. Early methods predominantly relied on depth sensors or RGB-D data to explicitly capture object geometry. For instance, methods like Point Pair Features (PPF) \cite{drost2010model} leverage depth maps to match geometric primitives such as point pairs against CAD models yet struggle under sensor noise and occlusions. With the rapid advancement of learning-based techniques, depth-based networks \cite{brachmann2016uncertainty} have integrated geometric priors into neural architectures to enhance pose estimation. Despite their effectiveness, the strong reliance on high-quality depth measurements limits their applicability in RGB-only settings.

\par A significant shift emerged with the development of multi-view RGB optimization frameworks. Early works such as \cite{labbe2020cosypose} introduced correspondence-free formulations that directly regress object poses by optimizing the alignment between rendered and observed images. Building upon this paradigm, \cite{yang2023icra} proposed a two-stage decoupled optimization strategy, where 3D translation is first resolved using scale-aware keypoints, followed by rotation estimation enforced through geometric consistency. Although this design alleviates depth ambiguity inherent in single-view RGB inputs, it relies heavily on accurate keypoint detection, which is often unreliable for textureless objects. Subsequent approaches, exemplified by OnePose++ \cite{he2022onepose++}, replaced handcrafted keypoints with detector-free feature matching techniques, enabling the reconstruction of sparse 3D point clouds from multi-view RGB sequences. While these methods improve generalization across objects and scenes, their reliance on sparse geometric representations frequently fails to resolve pose ambiguities in symmetric or planar regions due to geometric ambiguity. Recently, the paradigm has shifted toward neural implicit representations and explicit Gaussian splatting to bridge the gap between geometry and appearance. However, despite their high-fidelity rendering capability, these approaches still require stronger underlying geometric constraints to fully resolve the pose ambiguities inherent to textureless objects.

\subsection{Gaussian Splatting for Generalizable Pose Estimation}
\par In generalizable object pose estimation, the evolution of 3D object representation has followed multiple exploratory pathways. Traditional methods typically rely on pre-defined 3D CAD models~\cite{peng2019pvnet, Li_Wang_Ji_2019, Xiang_Schmidt_Narayanan_Fox_2018, ornek2024foundpose} and 3D point clouds~\cite{sun2022onepose, he2022onepose++} as object representations. The introduction of 3D Gaussian Splatting (3DGS)~\cite{kerbl20233d} revolutionized scene representation by offering explicit and high-fidelity rendering capabilities.

\par Recent efforts have adapted 3DGS for pose estimation tasks. GS-Pose~\cite{cai2024gs} was the first to leverage 3DGS for generalizable pose estimation, implicitly encoding objects as differentiable Gaussian distributions. However, this method relies on retrieval-based initialization, which often yields inaccurate initial poses. GS2Pose~\cite{mei2024gs2pose} extended this line of work by introducing a two-stage architecture. It employs a lightweight U-Net for coarse pose estimation and a GS-Refiner that utilizes Lie algebra to enable pose-differentiable rendering. A key innovation of GS2Pose is its ability to selectively update spherical harmonics and opacity during refinement, enhancing robustness against lighting variations and occlusions. Concurrently, iG-6DoF~\cite{cao2025ig} focuses on the initialization challenge for unseen objects. It constructs a multi-scale feature space using icosahedral group convolutions to generate robust pose hypotheses, which are then refined through an iterative render-and-compare strategy using 3DGS. Despite these advancements, existing 3DGS-based methods generally optimize Gaussian parameters primarily based on photometric loss. PG-Pose diverges from this trend by exploiting Planar-based Gaussian Splatting~\cite{chen2024pgsr}. We parameterize Gaussians with planar priors, enforcing coplanarity constraints and normal alignment during the reconstruction phase. This geometric regularization ensures that the 3DGS model maintains structural integrity, providing a more reliable foundation for subsequent pose optimization.

\section{PROPOSED FRAMEWORK}
\par In this section, we present PG-Pose, a robust 6D pose estimation framework for 6D pose estimation. The PG-Pose consists of two stages: offline representation extraction and online pose inference. In the offline phase, three distinct representations are obtained sequentially from unseen objects with known poses and are performed offline once per object. During inference, PG-Pose estimates the initial pose and the refined pose from the input image using the pre-built object representations in a cascaded manner. The overall architecture of the proposed framework is shown in Fig. \ref{arch}.

\subsection{Offline Representation Extraction}

\textbf{Semantic Representation.} Following GS-Pose \cite{cai2024gs}, we adopt the same semantic representation extraction pipeline. While this part is identical to GS-Pose, we retain it to ensure fair comparison and compatibility with subsequent modules. To establish robust semantic priors for textureless object understanding, we extract object semantic tokens ${F}_{obj}$ from multi-view reference images. We first select $N_{k}\left(\ll N_{r}\right)$ keyframes using farthest point sampling (FPS) \cite{he2022onepose++} from $N_{r}$ reference images. From these keyframes, we extract multi-view feature tokens ${F}_{fps} \in \mathbb{R}^{N_{k} \times L \times C}$, which are reshaped into spatial feature maps $\hat{{F}}_{fps} \in \mathbb{R}^{N_{k} \times L \times C}$, where L and C denote the token number and feature dimension of each frame. Next ${F}_{fps}$ and  $\hat{{F}}_{fps}$ are sent to a Co-segmenter module. Co-segmenter sample frame-wise center tokens $\hat{{F}}_{c}^{f p s} \in \mathbb{R}^{N_{k} \times C}$ from the 2D centroids of $\hat{{F}}_{fps} \in \mathbb{R}^{N_{k} \times L \times C}$ to encapsulate global object semantics. These center tokens guide a transformer-like module that processes ${F}_{fps}$ through $L_{m}$ stacked self- and cross-attention layers, The process can be formulated as
\begin{align}\label{1}
L_{m} \times\left\{\begin{array}{l}
{F}_{f p s}=\operatorname{Self} \operatorname{Attn}\left(F_{f p s}\right) \in \mathbb{R}^{N_{k} \times L \times C}  \\
{F}_{f p s}=\operatorname{Reshape}\left({F}_{f p s}\right) \in \mathbb{R}^{1 \times N_{k} L \times C} \\
{F}_{f p s} =\operatorname{CrossAttn}\left({F}_{f p s}, \hat{{F}}_{c}^{f p s}\right) \\
{F}_{f p s}=\operatorname{SelfAttn}\left({F}_{f p s}\right) \in \mathbb{R}^{1 \times N_{k} L \times C} \\
{F}_{f p s}=\operatorname{Reshape}\left({F}_{f p s}\right) \in \mathbb{R}^{N_{k} \times L \times C}
\end{array},\right.
\end{align}
where $L_{m}$ is the depth of the module. This hierarchical attention mechanism enhances intra-frame consistency and inter-frame alignment using the center tokens as semantic anchors. The refined tokens are then decoded by a mask consisting of two $3 \times 3$ convolutional layers and an upsampling layer to generate segmentation masks. Finally, ${F}_{obj}$ is extracted by masking $\hat{{F}}_{f p s}$ with the predicted regions, ensuring that only object-centric features are preserved for subsequent reconstruction.

\textbf{3D Object Gaussian Representation.}

\begin{center}
    \includegraphics[width=\columnwidth]{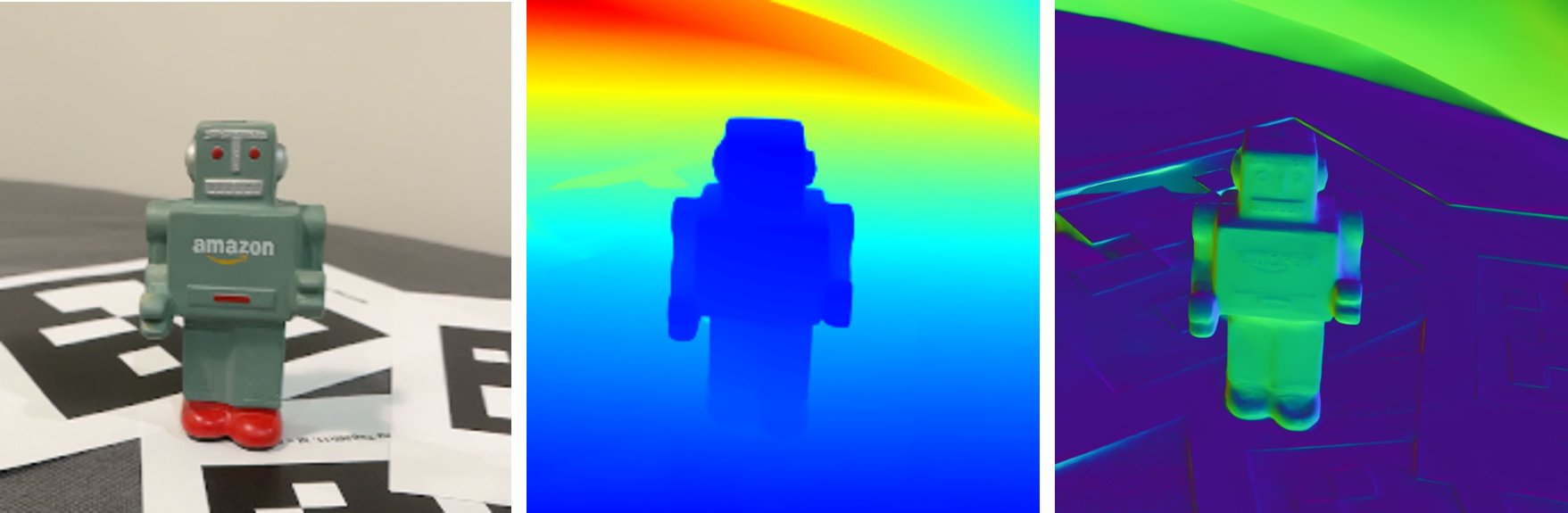}
    \captionof{figure}{Rendered depth and normal images by Planar-based Gaussian splatting}
    \label{pgsr}
\end{center}

Object segmentation is performed for all reference images using an Obj-Segmenter to obtain segmented reference images, which are utilized to build this 3DGO model. The PGS-based 3D object representation ${G}_{obj}$ is constructed by integrating planar geometric constraints into the 3D Gaussian Splatting framework \cite{kerbl20233d}, addressing the fragility of vanilla 3DGS in textureless regions. Inspired by planar-based GS \cite{chen2024pgsr}, we parameterize each planar-based Gaussian as ${G}_{obj} = \{\mu_i, r_i, s_i, \alpha_i, h_i\}_{i=1}^U$, where $\mu_i \in \mathbb{R}^3$ denotes the position, $r_i \in \mathbb{R}^4$ the rotation quaternion aligned with detected planar normals, $s_i \in \mathbb{R}^3$ the anisotropic scale, $\alpha_i \in \mathbb{R}$ the opacity, and $h_i \in \mathbb{R}^k$ the spherical harmonics coefficients.

Since planar normals cannot be directly obtained without an initial geometry, we adopt an iterative strategy to avoid circular dependency. First, an initial object model $G_{obj}^{(0)}$ is reconstructed using vanilla 3DGS without planar priors. Depth maps rendered from $G_{obj}^{(0)}$ are then back-projected from multiple viewpoints into a 3D point cloud, on which planar regions are detected using a plane fitting procedure. Although a point cloud could in principle be sampled directly from the Gaussian parameters, using back-projected depth maps yields view-consistent and visibility-aware 3D points by implicitly handling occlusions and filtering out Gaussians with low opacity or negligible contribution from a given viewpoint. This results in cleaner and more geometrically reliable point clouds for subsequent plane detection and orientation initialization. For each inlier plane, the normal vector $n_P$ is computed as the eigenvector corresponding to the smallest eigenvalue of the covariance matrix of the inlier points. Each Gaussian rotation $r_i$ is initialized to align explicitly with the closest detected plane normal. During optimization, coplanarity regularization further constrains Gaussians within the same plane to share consistent orientation. This process is repeated iteratively, alternating between plane detection and Gaussian refinement, until convergence is achieved. Each planar region corresponds to the inlier set of a detected plane obtained from the back-projected point cloud. As plane fitting is performed on spatially coherent neighborhoods, the resulting planar regions are mainly local. We do not explicitly enforce graph connectivity within a planar region. In this way, the construction of $G_{obj}$ becomes a well-defined refinement procedure rather than a one-step dependency, ensuring both logical consistency and stability in low-texture regions.

\textbf{3D Object Point Cloud Representation}
The 3D object point cloud ${P}_{o b j}$ is reconstructed through back-projection of depth maps rendered from the planar-based Gaussian representation ${G}_{o b j}$. Given a depth map $D$ of resolution $H \times W$, each pixel $(u, v)$ is transformed into a 3D point $ P_{i}  \in \mathbb{R}_{}^{3}$ by perspective projection: 
  \begin{equation}
    p_i = D(u,v) \cdot K^{-1}
    \begin{bmatrix}u & v & 1
    \end{bmatrix}^{\top},
  \end{equation}
where $K  \in \mathbb{R}_{}^{3\times 3}$ is the camera intrinsic matrix. 

\subsection{Online Pose Inference}

\begin{center}
    \includegraphics[width=\columnwidth]{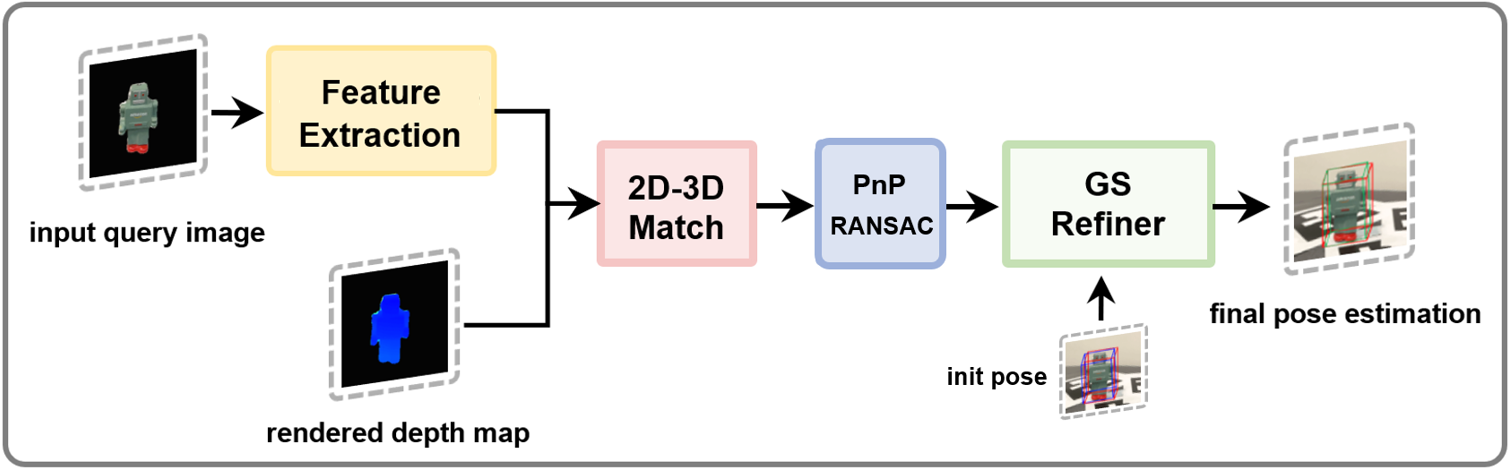}
    \captionof{figure}{The process of estimating the initial pose by pose estimator}
    \label{pnp}
\end{center}

During inference, the goal is to estimate the 6D pose of a query object image using the representations constructed in the offline stage. The process begins by cropping the target object from the input image with the help of the pre-computed semantic mask. From the cropped query image, sparse keypoints and descriptors are extracted using SuperPoint~\cite{detone2018superpoint}, a self-supervised keypoint detector and descriptor that has demonstrated robustness under low-texture conditions. Although SuperPoint is a learned feature extractor, it is class-agnostic and does not learn object-specific pose or geometry. In PG-Pose, it is only used to provide a sparse set of 2D-3D correspondences for Perspective-n-Point (PnP)-based pose initialization. In parallel, we render a set of RGB-D images from the planar-based Gaussian representation constructed offline. SuperPoint descriptors are computed on the rendered RGB images, and the corresponding depth maps are back-projected into 3D space to form a template point cloud, where each 3D point is associated with the descriptor of its 2D projection.

Keypoint descriptors from the query image are then matched to the template descriptors using Fast Approximate Nearest Neighbor (FLANN) \cite{muja2009fast}, establishing 2D-3D correspondences between the query image and the canonical 3D object representation. With these correspondences, an initial 6D pose is solved using a PnP algorithm with Random Sample Consensus (RANSAC) \cite{fischler1981random}, which ensures robustness against outliers. This initialization provides a coarse but geometrically consistent estimate of the object pose.

Finally, the initial pose is refined through the GS-Refiner, which performs differentiable rendering-based optimization. In this stage, the 3D Gaussian representation constructed offline remains fixed, while only the 6-DoF pose parameters are updated. The refiner minimizes photometric and geometric consistency losses between the rendered and observed images, thereby correcting residual errors from the PnP initialization. By focusing exclusively on pose refinement, this module ensures both computational efficiency and accurate convergence to the final pose estimate.

\section{EXPERIMENTS}
\subsection{Datasets and Evaluation Metrics}

The OnePose-LowTexture dataset is a standardized benchmark dataset specifically designed for the 6D pose estimation of low-texture objects. It comprises 40 everyday objects with low or no texture, covering typical scenarios in industrial sorting and domestic service robotics.

We evaluate pose estimation using ADD(S) that means the average distance between 3D points after being transformed by the ground truth and predicted poses. The formula for ADD(S) is defined as: 
\begin{equation}
 \operatorname{ADD}(\mathrm{S})=\frac{1}{N} \sum_{i=1}^{N} \min _{\mathbf{x}_{j} \in \mathcal{M}}\left\|\left(T \mathbf{x}_{i}-T^{\prime} \mathbf{x}_{j}\right)\right\| 
\end{equation}
where $N$ denotes the number of 3D model vertices, $T$ and $T^{\prime}$ represent the ground-truth and predicted rigid transformation matrices, ${x}_{i}$ is a 3D vertex from the object model, and $\mathcal{M}$ is the complete set of model vertices. The minimum operator addresses object symmetry by calculating the smallest Euclidean distance between a transformed vertex under the predicted pose $T^{\prime}$ and any geometrically equivalent counterpart in the symmetric model under $T$. This metric quantifies pose estimation accuracy as the average deviation across all vertices, with lower values indicating better alignment. The recall rate ADD(S)@0.1d further evaluates ADD(S) within 10\% of the object diameters.

\subsection{Training Details}

\par The proposed framework is trained in a unified manner to ensure both reliable reconstruction and accurate pose inference. In the offline stage, the planar-based Gaussian representation is optimized with a joint objective function consisting of photometric loss, normal alignment loss, and coplanarity regularization. The photometric loss enforces pixel-wise appearance consistency between the rendered image $I_{rend}$ and the ground-truth reference image $I_{ref}$:
\[
\mathcal{L}_{photo} = \frac{1}{|\Omega|}\sum_{p\in \Omega}\|I_{rend}(p)-I_{ref}(p)\|_2^2,
\]
where $\Omega$ denotes the foreground pixel set given by the segmentation mask. The normal alignment loss encourages geometric consistency by aligning the surface normal $n_{rend}(p)$ derived from the rendered depth with the Gaussian-inferred local normal $n_{gt}(p)$:
\[
\mathcal{L}_{norm} = \frac{1}{|\Omega|}\sum_{p\in\Omega}\big(1 - \langle n_{rend}(p), n_{gt}(p)\rangle\big).
\]
Finally, the coplanarity regularization penalizes deviations from detected planar regions. For each plane $P$, the distance of points $x\in P$ to the plane with normal $n_P$ and centroid $\bar{x}$ is minimized:
\[
\mathcal{L}_{plane} = \frac{1}{|P|}\sum_{x\in P}\big((x-\bar{x})^\top n_P\big)^2.
\]
For each planar region, the plane normal $n_P$ is estimated from the back-projected point cloud and serves as a geometric reference. For an individual Gaussian, its local surface normal is implicitly defined by its rotation parameter $r_i$. During optimization, the coplanarity regularization does not directly constrain the full rotation but instead enforces consistency between the Gaussian-inferred normal and the plane normal. The gradient of $L_{plane}$ is therefore propagated through the normal direction, which is a differentiable function of the rotation quaternion $r_i$, allowing the rotation to be updated via back-propagation. As a result, Gaussians assigned to the same planar region gradually align their normal directions while retaining flexibility in in-plane rotation.

The overall reconstruction loss is given by
\[
\mathcal{L}_{recon} = \lambda_{photo}\mathcal{L}_{photo} + \lambda_{norm}\mathcal{L}_{norm} + \lambda_{plane}\mathcal{L}_{plane},
\]
where $\lambda_{photo}, \lambda_{norm}, \lambda_{plane}$ are scalar weights balancing the objectives. The semantic segmentation network used to extract object masks is trained with a standard cross-entropy loss on annotated reference images. All networks and reconstruction components are optimized with the Adam optimizer, using an initial learning rate of $1\times10^{-3}$ and weight decay of $1\times10^{-4}$. 

\par During inference, the PGS-Refiner does not require any pre-training. Instead, it performs online optimization by minimizing a combination of photometric consistency, silhouette alignment, and geometric losses between the rendered and query images. Only the 6-DoF pose parameters are updated, while the Gaussian representation remains fixed.

 \begin{table}[ht]
    \setlength{\tabcolsep}{1.8pt}
    \centering
    \caption{6D pose estimation results on OnePose-LowTexture datasets. "init" indicates the initial pose estimation results}
    \label{tab:onepose_lowtexture}
    \resizebox{\columnwidth}{!}{%
    \begin{tabular}{*{10}{c}}
        \toprule
        Method & Toy. & Tea. & Cat. & Cam. & Shin. & Molie. & David & Marse. & Avg.\\
        \midrule
        Gen6D \cite{liu2022gen6d}
        & 55.5 & 40.0 & 70.0 & 42.2 & 62.7 & 16.6 & 15.8 & 8.1 & 38.9 \\
        OnePose \cite{sun2022onepose}
        & 65.6 & 89.0 & 39.7 & 90.9 & 87.9 & 31.2 & 42.7 & 30.4 & 59.7 \\
        OnePose++ \cite{he2022onepose++}
        & 89.5 & \textbf{99.1} & 97.2 & 92.6 & 98.5 & 79.5 & \textbf{97.2} & 57.6 & 88.9 \\
        \midrule
        GS-Pose \mbox{\hfill\tiny init} \cite{cai2024gs}
        & 55.0 & 75.7 & 82.6 & 69.7 & 95.1 & 63.4 & 65.7 & 57.5 & $70.6 \pm 0.7$ \\
        GS-Pose \cite{cai2024gs}
        & 89.3 & 86.7 & \textbf{100.0} & 90.2 & 99.3 & 95.9 & 91.7 & 83.6 & $92.1 \pm 0.3$ \\
        \midrule
        PG-Pose \mbox{\hfill\tiny init}
        & 67.5 & 77.1 & 79.3 & 72.4 & 96.2 & 73.5 & 71.2 & 64.2 & $75.1 \pm 0.6$ \\
        \textbf{PG-Pose}
        & \textbf{91.4} & 89.5 & 99.7 & \textbf{94.3} & \textbf{99.3} & \textbf{97.3} & 93.3 & \textbf{89.1} & $\mathbf{94.2 \pm 0.2}$ \\
        \bottomrule
    \end{tabular}%
    }
  \end{table}

\begin{table*}[t]
\centering
\caption{Ablation study on OnePose-LowTexture (ADD(S)@0.1d)}
\label{tab:ablation}
\begin{tabularx}{\textwidth}{cXXXXXXXXX}
\toprule
\textbf{Variant} & \textbf{Toy.} & \textbf{Tea.} & \textbf{Cat.} & \textbf{Cam.} & \textbf{Shin.} & \textbf{Molie.} & \textbf{David} & \textbf{Marse.} & \textbf{Avg.} \\
\midrule
w/o Planar Priors     & 88.6 & 86.7 & 96.1 & 89.2 & 96.5 & 92.3 & 90.4 & 82.5 & 90.8 \\
w/o Coplanarity Reg.  & 87.2 & 85.5 & 91.7 & 86.9 & 95.8 & 91.1 & 88.3 & 80.6 & 88.4 \\
w/o PnP Initialization& 83.1 & 82.6 & 89.3 & 84.5 & 93.4 & 87.2 & 86.5 & 78.1 & 85.6 \\
w/o GS-Refiner        & 79.3 & 80.7 & 85.6 & 81.2 & 91.5 & 84.1 & 83.7 & 75.9 & 82.7 \\
\textbf{PG-Pose (ours)} & \textbf{91.4} & \textbf{89.5} & \textbf{99.7} & \textbf{94.3} & \textbf{99.3} & \textbf{97.3} & \textbf{93.3} & \textbf{89.1} & \textbf{94.2} \\
\bottomrule
\end{tabularx}
\end{table*}

\subsection{Object Pose Estimation Performance Evaluation}

Table~\ref{tab:onepose_lowtexture} compares the 6D pose estimation accuracy of PG-Pose with prior methods on the OnePose-LowTexture dataset. Following the standard protocol, we report ADD(S)@0.1d for eight low-texture objects. For stochastic methods, including GS-Pose and PG-Pose, we repeat each experiment five times with different random seeds and report the mean and standard deviation. Overall, the proposed PG-Pose achieves the best performance across most categories, with an average ADD(S)@0.1d score of 94.2\%. Gen6D \cite{liu2022gen6d} relies on accurate 2D bounding box and template-matching, making it unreliable for textureless objects in this dataset. OnePose++ \cite{he2022onepose++} improves on OnePose \cite{sun2022onepose} by significantly improving the results to 88.9\%, but still does not exceed 90\%. GS-Pose has a large gap between the accuracy of its initial and refined poses due to the retrieval method used to obtain the initial pose.

\par In particular, the method demonstrates significant improvements on textureless and symmetric objects such as Cat, Cam, and Shin. For example, on the Cat object, PG-Pose improves the accuracy by 3.6\% over the baseline, while on Cam and Shin the improvements are 5.1\% and 4.8\%, respectively. These gains confirm the effectiveness of incorporating planar constraints and coplanarity regularization, which provide additional geometric cues when appearance information is unreliable. On relatively textured objects such as Toy and Tea, PG-Pose achieves comparable accuracy to the baseline while still delivering modest improvements, showing that the method does not sacrifice performance in cases where photometric information is already sufficient. Moreover, the robustness of the PnP-based initialization is reflected in the reduced variance of results across different categories, as the retrieved initial pose in prior methods often fails under heavy occlusion or symmetry. It is also worth noting that PG-Pose achieves superior performance on David and Marse, which are challenging due to partial texture and strong self-occlusion. The results suggest that the integration of planar priors with Gaussian refinement provides more reliable convergence, enabling accurate alignment even in difficult viewing conditions.

\begin{center}
    \includegraphics[width=\columnwidth]{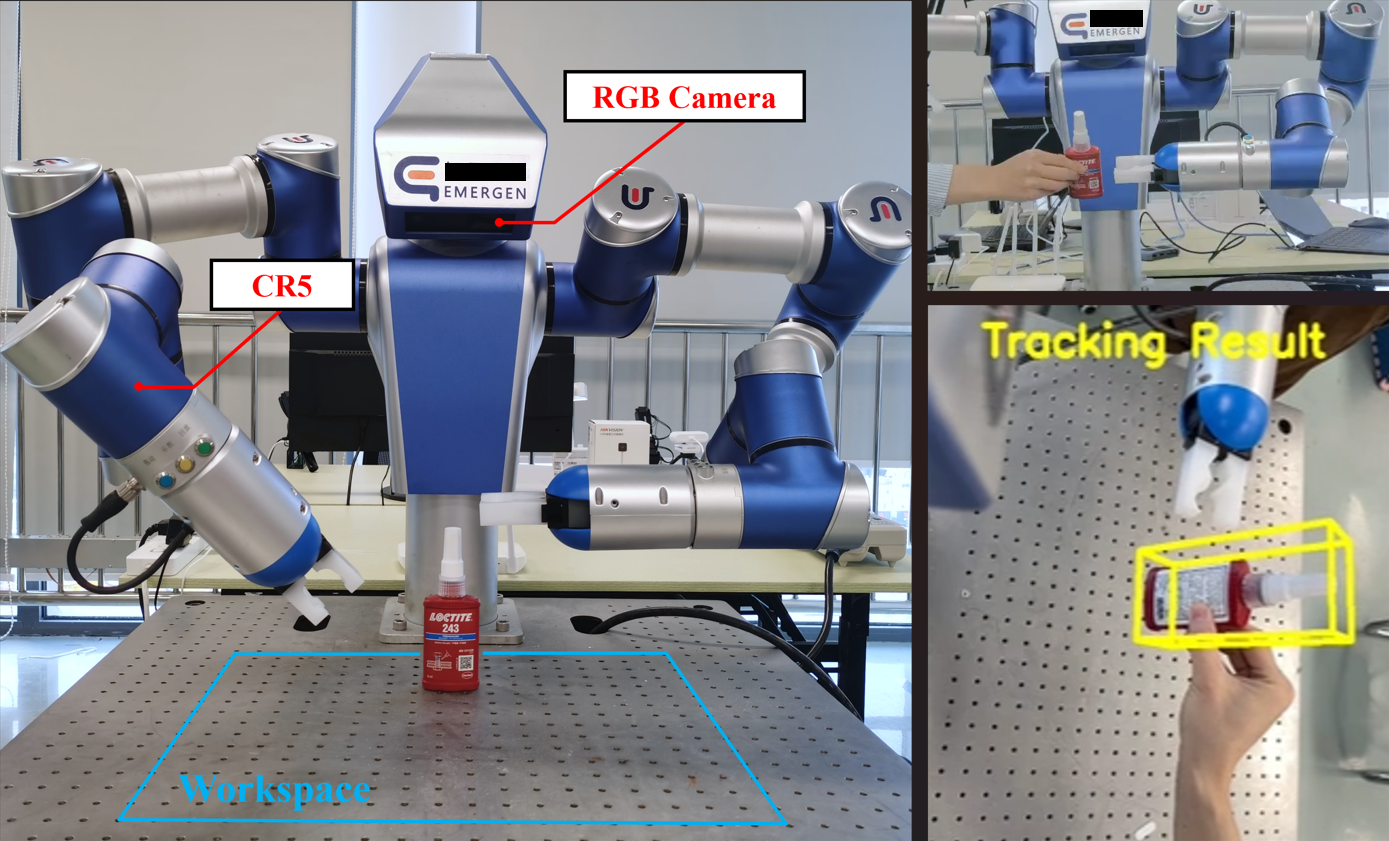}
    \captionof{figure}{Robotic experiments on the real-world grasping scenarios with PG-Pose.}
    \label{dual_arm}
\end{center}

\subsection{Real World Experiment on Grasp Tasks}

As shown in Fig.~\ref{dual_arm}, to validate the effectiveness of PG-Pose in real-world settings, we conduct robotic grasping experiments on an industrial collaborative dual-arm platform composed of two 6-DOF CR5 robotic arms, with an RGB camera mounted on the robot head.  To ensure a fair evaluation, we follow a standardized grasping protocol for all methods. The platform is equipped with a single parallel-jaw gripper, whose opening width is limited and whose contact stability is sensitive to the object geometry. Therefore, in this paper we restrict the real-world grasping evaluation to a single textureless object that is reliably graspable under our setup. The target object is an unseen plastic bottle placed randomly on a planar workspace without providing any CAD model or prior geometry. For each trial, the object pose is estimated from a single RGB image captured by the head camera, and a top-down parallel-jaw grasp is planned based on the predicted pose. A grasp is considered successful if the robot can hold it stably for more than 5\,seconds without slipping or collision. Each method is evaluated over 50 independent trials with random object placements. 

We use OnePose++ \cite{he2022onepose++} as the baseline method and the proposed PG-Pose for the tests. As a result, the manipulator successfully grasped the target objects with a success rate of 92.3\%, compared to 78.5\% for the baseline method. The results show that PG-Pose enables reliable grasp execution. The predicted poses remain consistent under partial occlusions, which are common in unstructured environments. We also observe that most failures are associated with unstable gripper contacts.

\subsection{Ablation Study}

To further clarify the contribution of each component in PG-Pose, we conduct ablation experiments on the OnePose-LowTexture dataset. Table \ref{tab:ablation} reports the results.

\textbf{w/o Planar Priors.} When removing planar constraints and only relying on photometric consistency as in GS-Pose, the average accuracy drops by 3.4\%, confirming that planar priors are crucial for textureless surfaces.

\textbf{w/o Coplanarity Regularization.} Removing coplanarity terms decreases performance especially on symmetric objects, showing the benefit of geometric structural constraints.

\textbf{w/o PnP Initialization.} Replacing PnP-based initialization with retrieval significantly reduces the accuracy of the initial pose, which propagates errors to the refinement stage.

\textbf{w/o GS-Refiner.} Without iterative refinement, the average accuracy drops to 82.7\%, indicating that Gaussian-based refinement is essential for final pose alignment.

\section{CONCLUSIONS}

\par This work presents PG-Pose, a geometry-aware approach for CAD-free 6D pose estimation of textureless objects. Rather than introducing a new pose estimation pipeline, PG-Pose enhances the geometric expressiveness of 3D Gaussian representations through planar priors, alleviating photometric ambiguity in low-texture scenarios and providing reliable geometric cues for pose initialization and refinement without CAD models. Experiments on the OnePose-LowTexture dataset demonstrate state-of-the-art performance, achieving 94.2\% ADD(S)@0.1d accuracy, with a 2.1\% improvement over the SOTA GS-based method. Moreover, real-world deployment on a dual-arm industrial robot validates the practical applicability of PG-Pose. Despite its effectiveness, PG-Pose has several limitations. The current representation primarily exploits locally planar geometric structures, which may be less optimal for objects dominated by smoothly curved surfaces such as spheres or highly cylindrical shapes. In addition, our real-world robotic validation is currently limited to a single representative object due to the constraints of the employed end-effector, leaving broader generalization to more diverse object categories and manipulation tasks for future exploration.

\bibliography{IEEEfull}

\end{document}